\documentclass[letterpaper, 10 pt, conference]{ieeeconf}  % Comment this line out if you need a4paper

\IEEEoverridecommandlockouts                              % This command is only needed if 
\usepackage{graphics} % for pdf, bitmapped graphics files
\usepackage{epsfig} % for postscript graphics files
\usepackage{times} % assumes new font selection scheme installed
\usepackage{amsmath} % assumes amsmath package installed
\usepackage{amssymb}  % assumes amsmath package installed
\usepackage{amsfonts}
\usepackage{textcomp}
\usepackage{cite}
\usepackage{fmtcount}
\usepackage[hidelinks]{hyperref}
\usepackage{url}
\usepackage{tabularx}
\usepackage{array}
\newcolumntype{Y}{>{\centering\arraybackslash}X}
\usepackage{float}
\usepackage{csquotes}
\usepackage{censor}

\title{\LARGE \bf
ColoACT: Multi-Cue Action Chunking for Smooth Autonomous Colon Navigation on a Self-Propelled Endoscopic Robot
}

\author{Jian Hu$^{1}$, Shujing He$^{1}$, Leixin Chang$^{2}$, Zongze Li$^{1}$, Ding Huang$^{1}$, \\Chaoyang Shi$^{3}$, and Chengzhi Hu$^{1}$, \IEEEmembership{Senior Member,~IEEE}% <-this % stops a space
\thanks{This work is supported by the National Key Technologies R\&D Program of China (Grant No. 2023YFC2415900), the National Natural Science Foundation of China (No. 62373182). This work is partly supported by the Shenzhen Science and Technology Program (Grant No. JCYJ20241202125417024), and Guangdong Basic and Applied Basic Research Foundation (Grant No. 2024A1515011915). The authors acknowledge the assistance of SUSTech Core Research Facilities.}% <-this % stops a space
\thanks{J. Hu, S. He, Z. Li, D. Huang and C. Hu are with the Department of Mechanical and Energy
Engineering, Southern University of Science and Technology, Shenzhen,
518055, China. (Corresponding email: hucz@sustech.edu.cn)}%
\thanks{L. Chang is with the School of Mechanical Engineering and ZJU-UIUC Institute, Zhejiang University, Zhejiang, China}%
\thanks{C. Shi is with the School of Mechanical Engineering, Tianjin University, Tianjin, 300072, China. }%
}

\begin{document}

\maketitle
\thispagestyle{empty}
\pagestyle{empty}

%%%%%%%%%%%%%%%%%%%%%%%%%%%%%%%%%%%%%%%%%%%%%%%%%%%%%%%%%%%%%%%%%%%%%%%%%%%%%%%%
\begin{abstract}

Autonomous colonoscopic navigation can reduce operator burden and the risk of loop formation or tissue trauma, but remains challenging due to deformable anatomy, weak-texture and specular endoscopic visuals, and contact-rich viscoelastic interactions. Existing methods either rely on geometry-driven pipelines, which are efficient and interpretable yet brittle due to manually engineered features and switching logic, or adopt learning-based policies, whose inferred depth/geometry can become temporally inconsistent or overly smooth under weak texture and specular highlights while simulation-trained variants (e.g., deep reinforcement learning) may further suffer from a sim-to-real gap. We propose ColoACT, an autonomous navigation system that integrates an RGB-D-E based Action Chunking Transformer policy (ColoACT policy) for a compact self-propelled Bevel-Gear-Based Endoscopic Robot (BGER). The ColoACT policy augments RGB with estimated relative depth and a gradient-based pseudo-elevation map to enhance fold-ridge saliency and other high-frequency geometric cues, and enables smooth continuous control of the BGER by predicting overlapping action chunks and fusing them via temporal ensembling. In different \textit{ex-vivo} porcine colons (approximately 60 cm), our system achieves success rates of 85.4\% and 72.5\% in straight and curved segments, respectively, and achieves 70\% success in 90-degree turns and 60\% in double-bend sequences, with feasibility further demonstrated in challenging triple-bend segments. The project page is available at: \url{https://Adamhu1.github.io/ColoACT/}.

\end{abstract}

%%%%%%%%%%%%%%%%%%%%%%%%%%%%%%%%%%%%%%%%%%%%%%%%%%%%%%%%%%%%%%%%%%%%%%%%%%%%%%%%
\section{INTRODUCTION}
Endoscopic examination remains the gold standard for the prevention and early diagnosis of colorectal diseases\cite{van2014conditional}. While traditional colonoscopy is widely adopted, it still suffers from a steep learning curve, limited automation, and risks of loop formation or tissue trauma during navigation\cite{alazmani2016quantitative}. Robotic endoscopic systems are therefore being explored to improve navigational safety and consistency\cite{mo2022task}, and to reduce the cognitive and physical burden on physicians.

Autonomous colonoscopic navigation is challenging due to large anatomical variability with frequent bends and collapses\cite{song2022novel}, weak texture and specular highlights, and strongly nonlinear, time-varying interactions\cite{duan2024operating} between the endoscope and viscoelastic tissue\cite{cui2023hybrid}. These factors demand a closed-loop policy that is robust under imperfect perception and contact disturbances while producing smooth, stable motion. Existing robotic navigation solutions fall into geometry-driven pipelines\cite{prendergast2018platform,prendergast2020real,prendergast2018autonomous} and learning-based approaches\cite{yip2023artificial,ozyoruk2021endoslam,he2024monolot}. Geometry-driven methods are often efficient and interpretable, but their reliance on manually engineered features and switching logic can make the behavior brittle under challenging visual conditions and contact-rich interactions. 
Learning-based strategies derive policies directly from data and can handle nonlinear dynamics. However, they remain limited by unreliable geometric cues under weak-texture/specular observations\cite{khan1996vision}, where learned depth/geometry can be temporally inconsistent or overly smooth and obscure fine topographic structures\cite{widya2021self}, and by sim-to-real gaps in simulation-trained variants (e.g., reinforcement learning) due to difficulty of faithfully modeling viscoelastic, contact-rich colon interactions\cite{bhattarai2022virgin,bharucha2001viscoelastic}.

To address the challenges of robustness under weak-texture conditions, smooth control consistency, and the sim-to-real gap, we propose ColoACT, a holistic navigation system that integrates an RGB-D-E input that combines appearance (RGB) and geometry (RGB with estimated relative depth and a gradient-based pseudo-elevation map) based Action Chunking Transformer (ACT) policy for smooth, continuous closed-loop control and a compact self-propelled Bevel-Gear-Based Endoscopic Robot (BGER) designed to provide the mechanical and visual stability required by learning-based navigation. To the best of our knowledge, we present the first imitation-learning-based navigation policy for a self-propelled colonoscopic robot in \textit{ex-vivo} tests. We validate the full system in \textit{ex-vivo} porcine colons, achieving success rates of 85.4\% in straight and 72.5\% in curved segments (\(\sim\)60 cm). The system further succeeds in highly tortuous anatomies, successfully negotiating sharp 90-degree turns (70\%) and double-bend paths (60\%), and remains feasible even in extreme triple-bend scenarios. The contributions of this work are as follows:

\begin{figure*}[!t]
\centerline{\includegraphics[width=\textwidth]{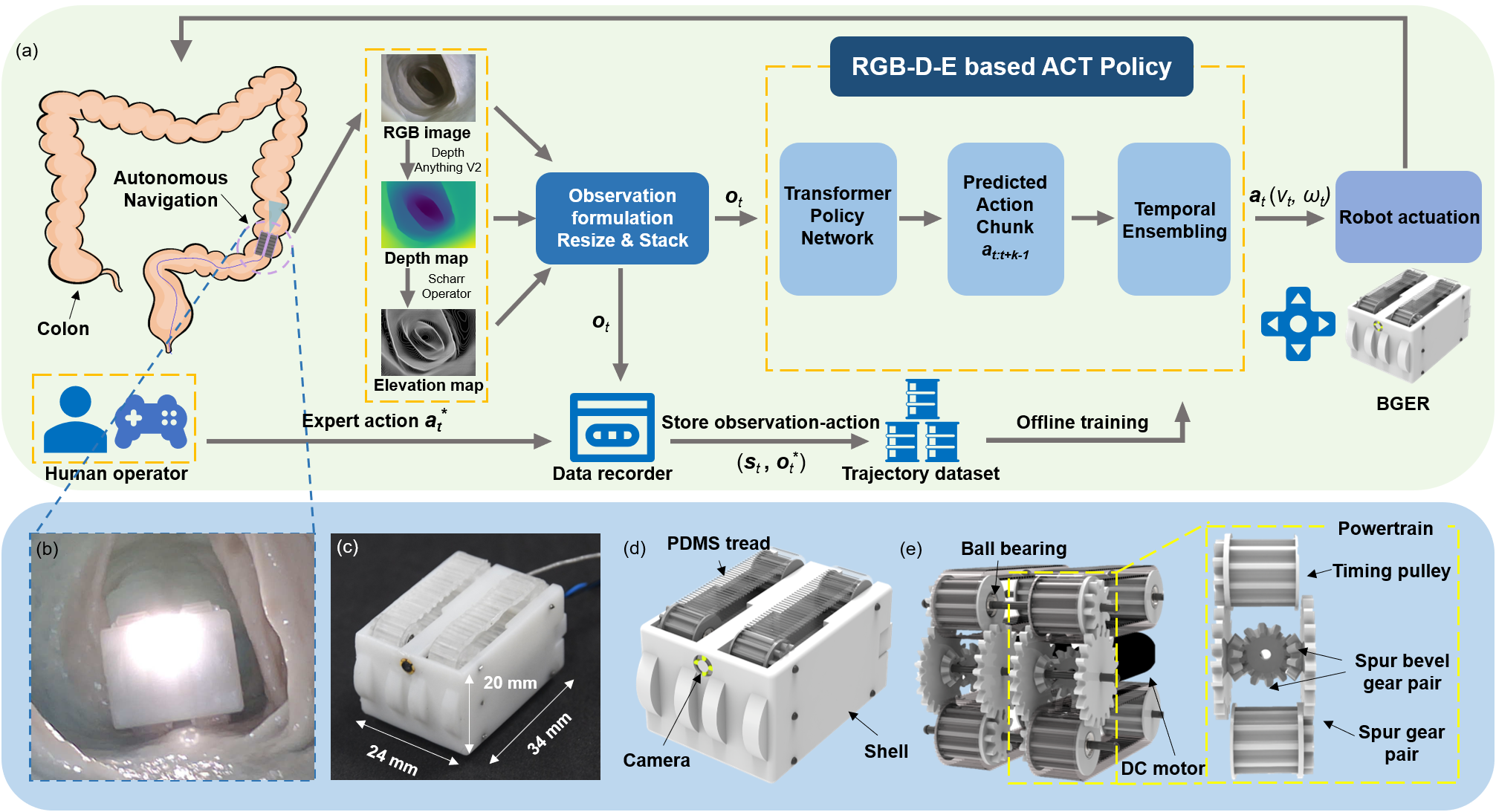}}
\caption{System Overview of the proposed ColoACT for Autonomous Navigation of the Bevel-Gear-Based Endoscopic Robot. (a) Framework Overview: The framework integrates a multi-cue perception module (RGB, relative depth, and pseudo-elevation, RGB-D-E) with an ACT-based control policy. Left: $\mathbf{o}_t = [I^{\mathrm{rgb}}_t, D_{t}, E_{t}]$ is formed using Depth Anything V2-Base for relative depth and Scharr gradients for pseudo-elevation. Right: The policy predicts an action chunk $\mathbf{a}_{t:t+k-1}$, which is synthesized via Temporal Ensembling to generate smooth actuation commands $\mathbf{a}_t$ ($v_t$,$\omega_t$). ($v_t$,$\omega_t$) represents the normalized linear velocity and angular velocity. (b) The actual navigation of the BGER in the porcine colon. (c) The assembled BGER, with dimensions of 24 × 20 × 34 mm. (d) The overall design of the BGER. (e) Transmission System Design of the BGER.}
\label{image_framework}
\end{figure*}

\begin{enumerate}
\item We proposed an RGB-D-E ACT navigation policy (ColoACT Policy) that leverages multi-cue geometric representations to improve robustness in challenging colonic environments, while generating smooth and stable actions. 
\item We design a compact self-propelled BGER mechanism that ensures transmission stability and supports reliable visual feedback for autonomous control. 
\item We demonstrate system-level validation of ColoACT on \textit{ex-vivo} porcine colons, showing strong performance across straight, curved, and complex tortuous segments.
\end{enumerate}

\section{Related work}
\subsection{Geometry-Driven Navigation for Robotic Endoscopy}
Geometry-driven pipelines have been widely investigated for autonomous navigation in the colon. A common paradigm is to extract explicit geometric features from endoscopic observations and then derive motion commands through deterministic logic. Typical approaches estimate the lumen direction/center by fitting geometric primitives (e.g., circles\cite{prendergast2018platform}, ellipses and triangles\cite{prendergast2020real}) or by detecting haustral folds\cite{he2024track} to guide the robot. These methods are attractive for interpretability and computational efficiency, but often require manually engineered thresholds and rule-based mode switching, leading to discrete behaviors and sensitivity to perception noise and contact-induced disturbances, especially under weak texture and specular highlights. Moreover, these studies are validated primarily on idealized silicone phantoms\cite{zhang2020enabling}, whereas real colons are deformable and time-varying, so the lumen geometry may deviate substantially from assumed analytic shape, thereby limiting generalization. Consequently, geometry-driven pipelines often fail to sustain stable and consistent closed-loop navigation when visual cues are unreliable or when the colon undergoes deformation and intermittent collapse.

\subsection{Learning-Based Navigation for Robotic Endoscopy}
Learning-based approaches for intraluminal navigation span geometry inference and policy learning. On perception, deep models estimate depth\cite{ozyoruk2021endoslam,he2024monolot} and mapping\cite{ma2021rnnslam,morlana2024colonmapper} to recover lumen geometry, but colon deformation, weak texture, and specularities often impair accuracy and temporal consistency\cite{khan1996vision}; moreover, monocular depth can over-smooth texture-poor regions, attenuating fine cues such as haustral-fold ridges that help infer lumen direction. On control, Reinforcement learning (RL) offers model-free nonlinear navigation with demonstrations in bronchoscopy\cite{zhao2024bronchocopilot} and colonoscopy\cite{tan2025safe,ng2024navigation} settings, yet safety constraints push training to simulation, where viscoelastic, contact-rich interactions are hard to model, yielding a pronounced sim-to-real gap. IL learns policies from expert demonstrations\cite{schmidgall2024robots}, reducing dependence on accurate physics\cite{ho2016generative}, but naive behavior cloning (BC) suffers a covariate shift and step-wise prediction can break long-horizon consistency\cite{ross2011reduction}, producing jittery maneuvers. These gaps motivate using explicit, stable geometric cues and temporally coherent closed-loop actions under visual degradation and contact disturbances.

\begin{figure*}[!t]
\centerline{\includegraphics[width=\textwidth]{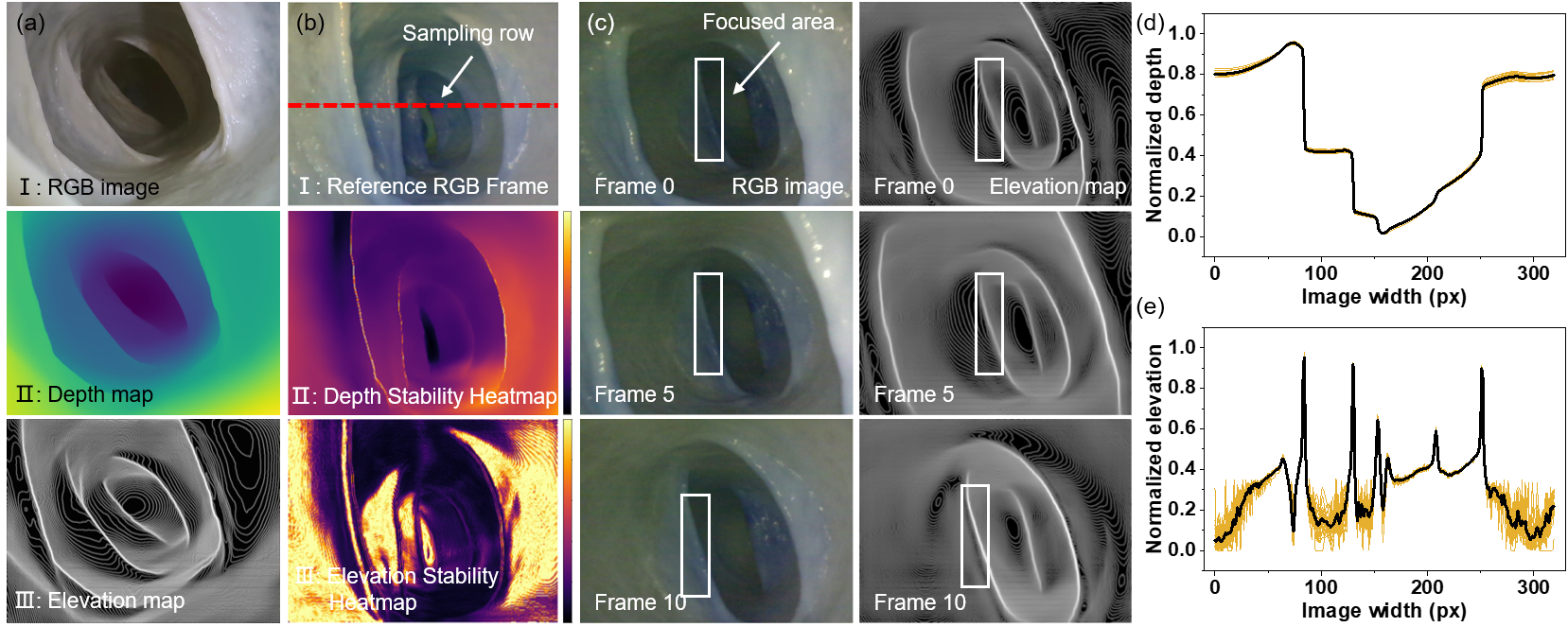}}
\caption{Visualization and Stability Analysis. (a) Multi-Cue inputs: RGB, relative depth, and pseudo-elevation map. (b) Temporal standard deviation heatmaps ($N=50$); the red line marks the sampling row. (c) Dynamic consistency sequence showing stable feature evolution. Cross-sectional profiles at the sampling row of (d) Depth and (e) Elevation. Black lines denote the mean signal, while yellow clusters show raw noise distribution, confirming spatial peak localization. }
\label{visual}
\end{figure*}

\section{Method}
As illustrated in Fig.~\ref{image_framework}(a), ColoACT performs closed-loop colon navigation by integrating multi-cue perception, an action-chunking policy, and online fusion of overlapping action chunks via Temporal Ensembling on the BGER platform.
In the perception phase, the system augments the monocular RGB input with a relative depth map (denoted as depth, $D$) and a gradient-based pseudo-elevation map (denoted as elevation, $E$) to form the observation $\mathbf{o}_t$. This explicit geometric encoding enhances feature extraction in weak-texture environments.
For control, the policy network predicts a sequence of future actions (action chunk $\mathbf{a}_{t:t+k-1}$) rather than a single step. These overlapping chunks are synthesized using Temporal Ensembling~\cite{Zhao-RSS-23} to generate smooth and continuous motor commands $\mathbf{a}_t = [v_t, \omega_t]^\top$ for the BGER, where $v_t$ and $\omega_t$ represent normalized linear and angular velocities, respectively. This mechanism effectively filters high-frequency jitter and ensures navigational stability.

\subsection{ColoACT Policy: Multi-Cue-Based ACT Policy for Colonoscopic Navigation via Human Demonstration}

\subsubsection{Multi-Cue Visual Input Information Processing} To perceive the 3D structure of the weak texture colon environment, we construct a Multi-Cue observation space, $\mathbf{o}_t = [I^{\mathrm{rgb}}_t, D_{t}, E_{t}]$. We leverage the pre-trained Depth Anything V2-Base model \cite{yang2024depth} as a robust depth estimator that infers dense relative depth maps (\text{${D}_{raw}$}) directly from single RGB frames.

To standardize the policy network input, the raw depth values are normalized to $[0, 1]$. Furthermore, since raw depth maps often exhibit local smoothing, we introduce a gradient-based pseudo-elevation map (${E}$) to explicitly encode high-frequency topographic features such as folds and fine-scale relief. This is computed by applying the Scharr gradient operator to the normalized depth map ${D}$:
\begin{equation}
M=\|\nabla{D}\|_2
\end{equation}
\begin{equation} 
E = \frac{L - \min(L)}{\max(L) - \min(L)}
\end{equation}
where $\|\nabla {D}\|_2$ denotes the gradient magnitude and the logarithmic transformation compresses the dynamic range of strong gradients while amplifying weak but informative boundaries, making fold-ridge cues more distinguishable, and \text{$L = \ln(1+M)$}; the result is then min–max normalized to $[0, 1]$. The final observation $\mathbf{o}_t$ is formed by concatenating the RGB, depth, and elevation maps, combining semantic context with explicit geometric constraints. A set of generated RGB, depth, and elevation maps is shown in Fig. \ref{visual}(a). As shown in Fig. \ref{visual}(a)-III, the elevation map explicitly encodes the high-frequency ridges of the colon folds and the subtle topographic relief, which appear merely as subtle slopes in the depth map (Fig. \ref{visual}(a)-II). Raw RGB images were captured at $640 \times 480$ resolution to ensure clarity for operator, but were resized to $320 \times 240$ before being fed into the ResNet-18 backbone to optimize computational efficiency.

To ensure robust control, we evaluated temporal consistency over $N=50$ static frames. As shown in Fig. \ref{visual}(b), the resulting heatmaps show low pixel-wise standard deviations for depth ($\sigma=0.0071$) and elevation ($\sigma=0.0250$), and as shown in Fig. \ref{visual}(d-e), cross-sectional analysis along the sampling row confirms that despite minor gradient noise, the spatial peaks representing tissue folds remain localized without shifting. In addition, we simultaneously evaluated the dynamic stability of the visual perception module. As shown in Fig. \ref{visual}(c), as the BGER approaches the fold from frame 0 to 10, the geometric structure marked in the white square in the elevation map evolves consistently with the visual structures in the RGB image, with clear boundaries and no artifacts. This suggests that our perception module extracts physically meaningful geometric cues rather than transient noise.

\subsubsection{Predictive Action Chunking for Smooth Endoscopic Navigation}
To achieve continuous motion control of the BGER and handle non-Markovian dynamics and hysteresis effects in viscoelastic environments, we propose multi-cue endoscopic observation $\mathbf{o}_t$ on top of the Action Chunking with Transformers (ACT)\cite{Zhao-RSS-23}, a generative policy framework based on conditional variational autoencoders (CVAE), which can predict future action sequences to ensure temporal coherence of navigation.

Let observation at time $t$ be $\mathbf{o}_t = [I^{\mathrm{rgb}}_t, D_{t}, E_{t}] \in \mathbb{R}^{H \times W \times C}$, where $C=5$ represents the concatenated channels. The action space is $\mathbf{a}_t = [v_t, \omega_t]^\top$, representing normalized linear and angular velocities. At each time step, the policy predicts a chunk of actions $\mathbf{a}_{t+k-1} = \{a_t, a_{t+1}, \dots, a_{t+k-1}\}$ with a prediction horizon $k=16$.

The visual observation $\mathbf{o}_t$ is first encoded into a feature embedding $e_{vis}$ via a modified ResNet-18 backbone $f_{\theta}$:
\begin{equation}
e_{vis} = f_{\theta}(\mathbf{o}_t), \quad e_{vis} \in \mathbb{R}^{d}
\end{equation}
where $d=256$. To capture the stochasticity of human demonstrations, the CVAE encoder approximates the posterior distribution $q_\phi(z | \mathbf{o}_t, \mathbf{a}_{t+k-1})$ conditioned on the current observation and the ground-truth action chunk. A latent variable $z$ is sampled from a diagonal Gaussian distribution parameterized by the encoder outputs $\mu$ and $\sigma$:
\begin{equation}
z = \mu + \sigma \odot \epsilon, \quad \epsilon \sim \mathcal{N}(0, I)
\end{equation}

The decoder functions as the control policy $\pi_\psi$, generating the action chunk conditioned on the latent style $z$ and visual context $e_{vis}$:
\begin{equation}
    \hat{\mathbf{a}}_{t+k-1} = \pi_\psi(z, e_{vis}, P)
\end{equation}
where $P \in \mathbb{R}^{k \times d}$ are fixed sinusoidal position queries that guide the ordering of the predicted sequence. 
The network is trained by minimizing the Evidence Lower Bound (ELBO):
\begin{equation}
    \mathcal{L} \!= \!\sum_{i=0}^{k-1} || \hat{\mathbf{a}}_{t+i} - \mathbf{a}_{t+i} ||^2_2\! + \!\beta \! D_{KL}\left(q_\phi({z}|\mathbf{o}_t, \mathbf{a}_{t:t+k-1}) || \mathcal{N}(0, I)\right)
\end{equation}

To mitigate jitter in the viscoelastic environment, we employ Temporal Ensembling. Since the policy predicts a horizon of k = 16 steps at each inference step, there are multiple overlapping predictions for any given time step $t$. We synthesize the final executed action $a_t$ using an ensemble average (Boxcar filter) of these overlapping chunks:
\begin{equation}
\mathbf{a}_t = \frac{1}{N_t} \sum_{i=0}^{N_t-1} \hat{\mathbf{a}}_{t}^{(t-i)}
\end{equation}
where $\hat{\mathbf{a}}_{t}^{(t-i)}$ denotes the action for time $t$ predicted at a previous step $t-i$, and $N_t$ is the number of valid overlapping predictions (typically $N_t=k$ in steady state). This mechanism acts as a low-pass filter, effectively smoothing out high-frequency noise from the generative policy and ensuring kinematic continuity essential for endoscopic safety.

\subsection{ColoACT Platform: System and Mechanical Design}
In this section, we elaborate on the mechanical design and the system integration of our proposed hardware platform, BGER.
The diameters of adult colons are approximately 3–8 cm overall\cite{alazmani2016quantitative}. To facilitate flexible maneuverability within the confined colon lumen, we developed a compact self-propelled tread robot with ultra-small dimensions of 24 × 20 × 34 mm (Fig.  \ref{image_framework}(c)). As shown in Fig. \ref{image_framework}(d), traction is provided by biocompatible PDMS treads featuring concave-convex textures to optimize friction against the intestinal mucosa\cite{he2024track}. The locomotion system is powered by two independent coreless DC micro-motors with reduction ratio of 700:1. To ensure stable movement and visual feedback for the learning policy, we devised a custom self-propelled Bevel-Gear-Based mechanism as shown in Fig. \ref{image_framework}(e). This configuration vectorially cancels radial and tangential reaction forces, effectively isolating the motor shafts from bending loads and reducing mechanical jitter. 
The BGER perceives the environment through a distal monocular vision module with a 3.3 mm diameter and an 85° FOV that supports video capture up to 720p at 30 fps to ensure sufficient spatiotemporal resolution. Intra-luminal illumination is provided by an integrated adjustable LED. A flexible tether transmits video data to the host computer, while an embedded microcontroller (Arduino Uno) generates PWM signals for low-level motor control based on the high-level policy commands. For safety, an emergency-stop button is provided to immediately disable motor actuation.

\section{Experiment}
Our goal in this section is to validate that the ColoACT policy, which combines multi-cue RGB-D-E observations with action chunking and temporal ensembling, enables robust and smooth closed-loop navigation of the BGER in visually degraded, contact-rich colonic environments. 
We evaluate ColoACT in zero-shot \textit{ex-vivo} deployments on unseen porcine colons, covering straight/curved and tortuous anatomies, and comparing against input-cue ablations and an RGB-D-E BC baseline. We then report offline ablations on a held-out set to quantify control stability and long-horizon prediction fidelity (ADE/FDE), complemented by open-loop rollouts.
\subsubsection{Training Setting}
We collected 214 teleoperated trajectories from \textit{ex-vivo} porcine colons, covering diverse scenarios including straight, turning, recovery, and visual artifacts (e.g., occlusions) to ensure robustness. The model was trained for 100 epochs with a batch size of 64. We utilized the AdamW optimizer with a learning rate of $10^{-4}$ and a weight decay of $10^{-4}$. The objective function follows the original ACT implementation, defined as a weighted sum of the reconstruction loss (MSE) and the KL-divergence loss, with the KL-weight $\beta$ set to 10.0. Training and offline evaluation were conducted on a single NVIDIA A100 GPU, while real-time deployment ran on an NVIDIA RTX 4060 GPU at 15 Hz.

\subsubsection{Ex-Vivo Evaluation of Autonomous Navigation in Porcine Colon}
Due to the high anatomical resemblance between porcine and human organs, particularly the presence of haustra and teniae coli in the colon \cite{similiarwithpigandhuman2012porcinemodel}, the pig serves as an ideal model for endoscopic research. Unlike silicone phantoms, real tissue preserves the complex nonlinear and viscoelastic characteristics inherent to the intestinal mucosa\cite{howard2021biochemical}. Therefore, we selected excised porcine large intestine for the \textit{ex-vivo} validation of the BGER.

\begin{table}[t]
\caption{Success rate of the ex-vivo experiment (\%)}
\label{success_rate_table}
\centering
\begin{tabularx}{\columnwidth}{|c|Y|Y|Y|Y|}
\hline
  & BC Policy& \multicolumn{3}{c|}{ACT policy} \\
\hline
Input mode  & RGB-D-E & RGB & RGB-D & RGB-D-E \\
\hline
Straight & 45.8 & 35.4 & 58.3 & 85.4\\
\hline
Curved & 57.5 & 25.0 & 50.0 & 72.5\\
\hline
\end{tabularx}
\end{table}

\begin{table}[t]
\caption{The normalized angular Jerk of RGB-D-E based BC policy and ColoACT policy in both straight and curved porcine colon ($a.u.$)}
\label{Jerk_table}
\centering
\begin{tabularx}{\columnwidth}{|c|Y|Y|}
\hline  
 & RGB-D-E based BC & RGB-D-E based ACT \\
\hline
Straight & 63.89 $\pm$ 7.8877& 4.74 $\pm$ 0.9312\\
\hline
Curve & 58.91  $\pm$ 6.5185& 4.47 $\pm$ 0.4298\\
\hline
\end{tabularx}
\end{table}

\begin{figure}[!t]
\centerline{\includegraphics[width=\columnwidth]{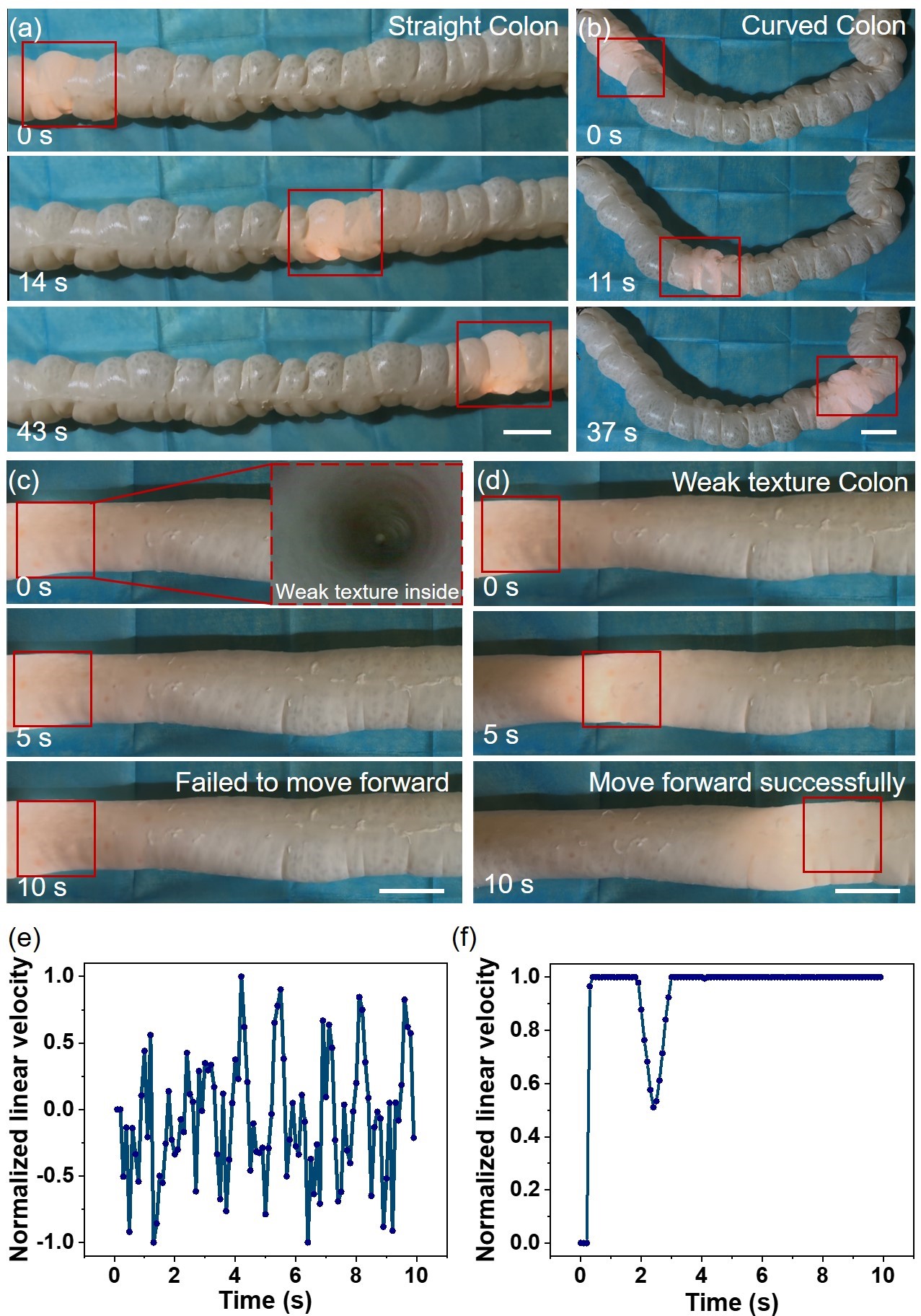}}
\caption{Validation of autonomous navigation in \textit{ex-vivo} porcine colon. Experiment in (a) Straight colon and (b) Curved colon. The movement of the BGER under weak texture conditions using (c) RGB-D-E based BC and (d) ColoACT policy. Inset photo shows the inside texture of the colon, Scale bar: 5 mm. The normalized linear velocity of (e) RGB-D-E based BC policy and (f) ColoACT policy when facing the weak texture.}
\label{experiment1}
\end{figure}

To validate the zero-shot generalization of our policy, we conducted deployment experiments on four novel porcine colons (\(\sim\)60 cm) excluded from the training set. Initially, we evaluated performance on fundamental navigation tasks, including straight and curved segments, as shown in Fig. \ref{experiment1}(a-b). The protocol involved 48 trials for straight sections and 40 for curved sections.
Under identical experimental conditions, the RGB-only ACT variant achieved success rates of 35.4\% in straight colon and 25.0\% in curved colon. Adding depth as an additional cue, the RGB-D ACT variant improved these rates to 58.3\% and 50.0\%, respectively. However, as listed in Table \ref{success_rate_table}, the proposed RGB-D-E ACT variant achieved the best performance among the tested input configurations, with success rates of 85.4\% and 72.5\%. This improvement is attributed to the complementary multi-cue inputs: while depth maps supply relative distance information, the elevation (pseudo-elevation) map explicitly encodes local surface curvature and gradient dynamics, allowing the policy to effectively distinguish the lumen center (target) from the colon wall (obstacles).

To evaluate the contribution of temporal action modeling beyond richer perception, we fix the input to RGB-D-E and compare ColoACT against a non-temporal BC baseline. Under the same conditions, the BC baseline achieved markedly lower success rates, with 45.8\% in straight and 57.5\% in curved, due to its greedy, non-temporal predictions, resulting in two representative failure modes. First, as listed in Table \ref{Jerk_table}, BC exhibits severe control instability. We use the normalized angular jerk on the commanded angular velocity $\omega$ to quantify steering smoothness over all trials. BC produces pronounced high-frequency oscillations in both straight and curved cases ($J_\omega$ $\approx 60$, $a.u.$) whereas our ACT-based policy, benefiting from action chunking and temporal ensembling, maintains substantially smoother steering ($J_\omega$ $<5$, $a.u.$), yielding an order-of-magnitude reduction in jitter. Second, BC frequently becomes disoriented in feature-poor regions (Inset photo of Fig. \ref{experiment1}(c)), as reflected by the erratic trajectory in Fig. \ref{experiment1}(c) and the fluctuating normalized linear-velocity profile in Fig. \ref{experiment1}(e). In contrast, our ColoACT policy exhibits smooth and continuous motion as illustrated in Fig. \ref{experiment1}(d), (f).

\begin{figure}[!t]
\centerline{\includegraphics[width=\columnwidth]{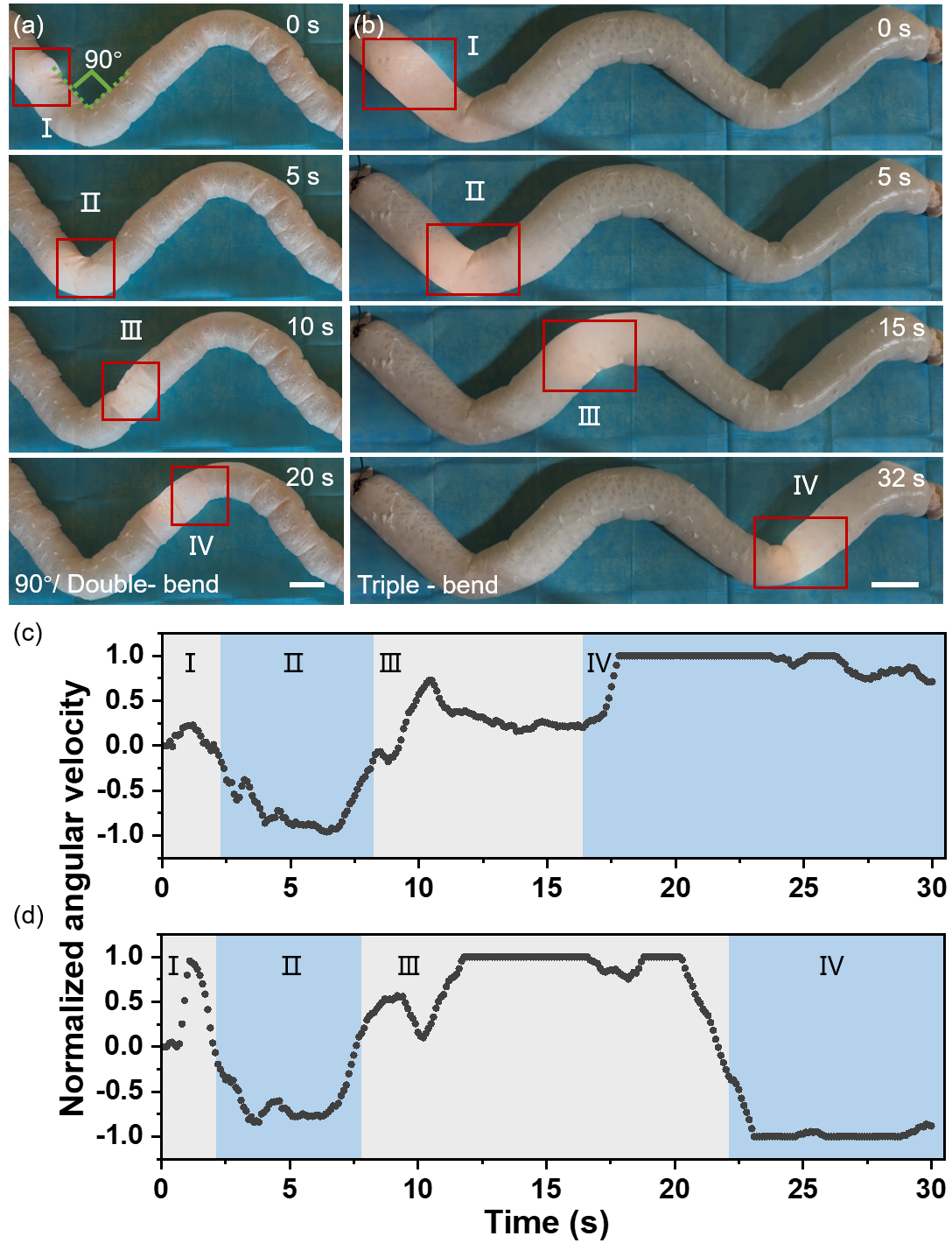}}
\caption{Experiment of ColoACT in complex colon, (a) 90° turns / double-bend and (b) triple-bend, Scale bar: 5 mm. The normalized angular velocity of the following (c) 90° turns / Double-bend and (d) Triple-bend colons.}
\label{experiment2}
\end{figure}

Finally, to validate the system's applicability to the complex morphology of the human-like colon, we tested the robot in highly tortuous scenarios involving 90-degree sharp turns, consecutive double bends, with two consecutive bends of \(\sim\)100° over \(\sim\)50 cm in total, and extreme triple bends, with three consecutive bends of \(\sim\)105° over \(\sim\)65 cm in total length as shown in Fig. \ref{experiment2}(a-b) and Supplementary Video. Normalized angular velocity profiles shown in Fig. \ref{experiment2}(c-d) exhibit a strong temporal correlation with the physical curvature of the \textit{ex-vivo} colon, divided into four distinct navigation phases (I–IV).

In the double-bend sequence in Fig. \ref{experiment2}(a), the control output in Fig. \ref{experiment2}(c) clearly mirrors the S-shaped geometry. After the initial entry (Zone I), the policy executes a sharp turn (Zone II, negative trough) followed by a smooth transition (Zone III). Crucially, in Zone IV, the angular velocity enters a positive saturation state, indicating the policy effectively locked its steering to navigate the prolonged second bend. Furthermore, in the highly challenging triple-bend scenario shown in Fig. \ref{experiment2}(b), the policy demonstrates adaptive behavior. The trajectory reveals a multi-stage polarity reversal pattern in Fig. \ref{experiment2}(d): transitioning from the first bend (Zone II, negative) to the second (Zone III, positive), and finally reversing again to tackle the third bend (Zone IV, negative). This continuous, oscillation-free switching confirms that the ACT-based policy successfully encodes long-horizon geometric context to handle continuous tortuous paths.

Our system demonstrated strong adaptability in complex maneuvers, achieving success rates of 70\% (7/10) in 90-degree turns and 60\% (6/10) in double-bend sequences. Furthermore, we explored the system's limits in triple-bend anatomies, achieving a feasible passage rate of 30\% (3/10). These results confirm that our proposed ColoACT policy can effectively negotiate intricate geometric constraints, providing a foundation for future clinical applications.

\subsubsection{Ablations on Temporal Ensembling and Input Cues, with Long-Horizon Open-Loop Rollout Validation}

\begin{figure}[t]
\centerline{\includegraphics[width=\columnwidth]{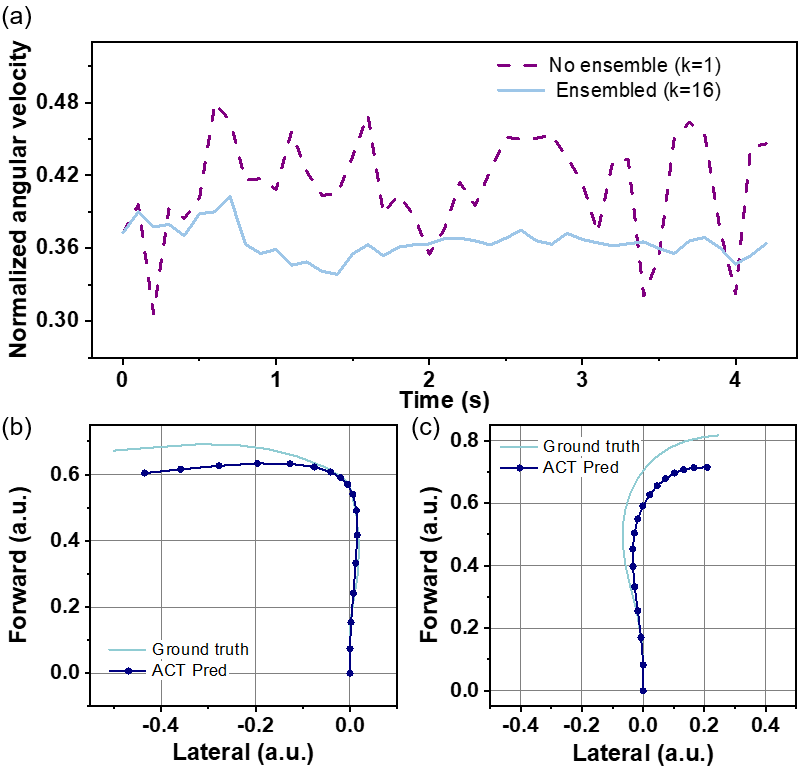}}
\caption{Temporal ensembling ablation and long-horizon open-loop rollout validation. (a) Temporal ensembling ($k=16$, blue) suppresses jitter compared to the baseline ($k=1$, purple). (b-c) Open-loop trajectory rollouts comparing ACT predictions (dark blue) with demonstrations (Ground truth, light blue) in normalized kinematic space.}
\label{ACTabout}
\end{figure}

To evaluate the factors behind the observed navigation performance, we perform offline ablations on temporal ensembling and input cues, and validate long-horizon behavior with open-loop rollouts. We first examine the impact of temporal ensembling on control stability across the validation set ($N=43$).  As shown in Fig. \ref{ACTabout}(a), the raw prediction without ensembling ($k=1$, purple dashed line) exhibits significant high-frequency oscillation, which risks motor service life and video stability. In contrast, by aggregating predictions over a sliding window ($k=16$), the ACT policy (solid blue line) effectively filters out stochastic noise. This generates smooth, continuous angular velocity commands essential for safe endoscopic maneuvers.

\begin{table}[t]
\caption{Open-loop trajectory prediction accuracy (ADE/FDE, $a.u.$) of the ACT policy across different input configurations; values are means over N = 43 validation samples.}
\label{ADE_table}
\centering
\begin{tabularx}{\columnwidth}{|c|Y|Y|Y|}
\hline
 & RGB & RGB-D & RGB-D-E \\
\hline
ADE (a.u.) & 0.1187 & 0.1096 & 0.0835\\
\hline
FDE (a.u.) & 0.2153 & 0.2076 & 0.1689\\
\hline
\end{tabularx}
\end{table}

To quantify long-horizon prediction fidelity under complex colon geometry, we employ the metrics of \textit{Average Displacement Errors} (ADE) and \textit{Final Displacement Errors} (FDE). Lacking odometry, these metrics are computed in the normalized kinematic space via open-loop integration of velocity predictions ($\mathbf{a}_{t:t+k-1}$) with $\Delta t = 1/15\ \mathrm{s}$. ADE averages the deviation over the full horizon ($k=16$), while FDE measures the deviation at the final step ($t+k-1$). Compared to the RGB-D baseline, the integration of elevation maps (RGB-D-E) reduced the ADE by 23.8\%, as shown in Table \ref{ADE_table}, indicating that the model maintains tight alignment with the expert's geometric intent. Furthermore, the FDE was reduced by 18.6\%, proving that explicit geometric encoding allows the model to infer the colon's high-frequency geometric cues far ahead, avoiding \enquote{short-sighted} planning errors, and also implies a stable prediction \enquote*{tail}, which is a prerequisite for the low-jerk control achieved via temporal ensembling.

Qualitative open-loop rollouts, as illustrated in Fig. \ref{ACTabout}(b-c), focus on single-instance chunks due to absent localization. These rollouts demonstrate that the predictions faithfully capture the underlying curvature of the expert's path rather than merely memorizing coordinates as well as the adaptability of the algorithm to the colon environment. Crucially, as the policy operates as a receding horizon controller, these long-term predictions serve as stable geometric priors, ensuring robust closed-loop tracking despite the inherent open-loop drift.

\begin{figure}[!t]
\centerline{\includegraphics[width=\columnwidth]{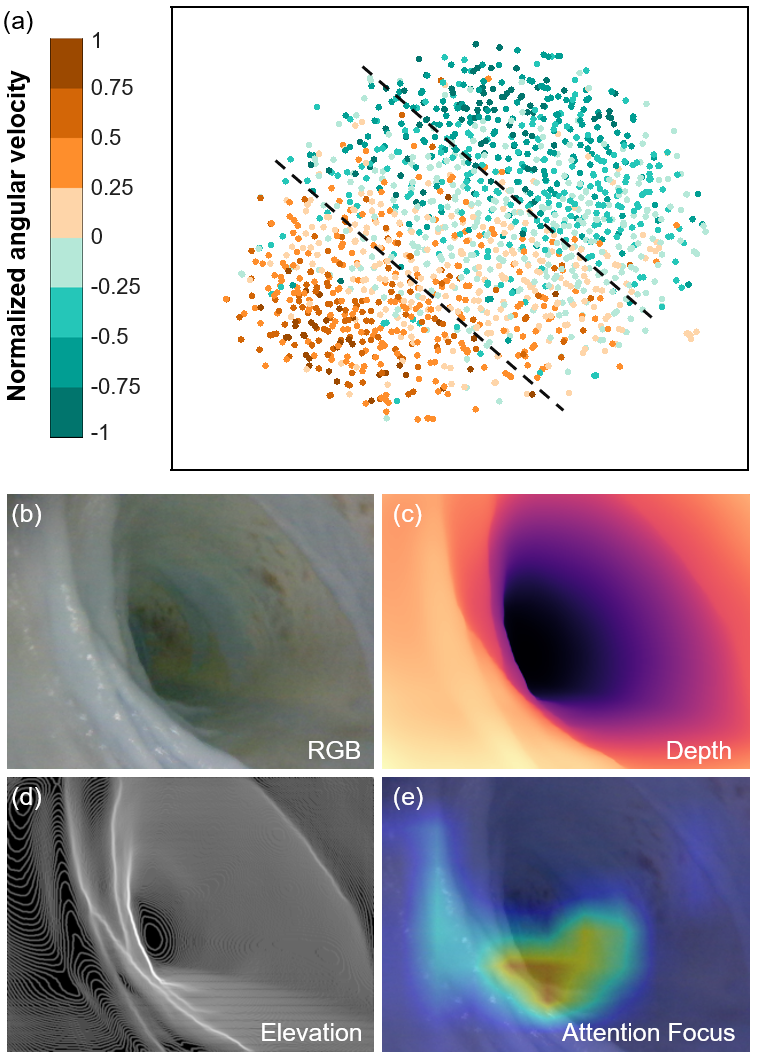}}
\caption{Semantic and attention analysis of policy behavior. (a) t-SNE visualization of the latent space, colored by normalized angular velocity, showing a structured and continuous manifold aligned with steering intent.  (b-d) Example observations: RGB image, depth map, and pseudo-elevation cue. (e) Grad-CAM visualization of the visual backbone highlights attention concentrated on the navigable lumen region and relevant anatomical boundaries during turning.}
\label{attention}
\end{figure}

\subsubsection{Semantic and Attention Analysis of Policy Behavior}

To investigate whether the model learns meaningful motion semantics rather than simply overfitting to training samples, we visualized the latent variable $z$ using t-SNE dimensionality reduction, as shown in Fig. \ref{attention}(a). The data points are color-coded by the normalized angular velocity of the associated action. The visualization reveals a highly structured manifold organization within the latent space. We observe a clear and continuous gradient transitioning from negative angular velocities (teal/green clusters) to positive ones (orange/brown clusters). This distinct clustering by steering intent suggests that the CVAE encoder has successfully disentangled the underlying geometric features of the navigation task. The smoothness of this latent manifold provides a robust basis for the generator to output continuous and stable motion controls, even in unseen scenarios.

To interpret the decision-making process of the ColoACT policy and further confirm that our method successfully identifies key visual features for navigation, we visualized the spatial attention of the visual backbone using Gradient-weighted Class Activation Mapping (Grad-CAM). Fig. \ref{attention}(e) illustrates the input-space attention mapping during a left-turning maneuver. Interestingly, rather than solely fixating on obstacles, the policy demonstrates a sophisticated, target-driven pathfinding behavior, the primary visual focus, highlighted in deep red, precisely anchors onto the distant, navigable lumen located on the left side, confirming the model's intent to steer towards safe free space. Concurrently, secondary attention (lighter activation) distributes along the left haustral fold ridges extracted by the elevation map shown in Fig. \ref{attention}(d). This attention pattern indicates that, guided by the multi-cue RGB-D-E input, the ColoACT policy effectively identifies the open trajectory while simultaneously maintaining awareness of adjacent topological boundaries, enabling smooth and collision-free navigation. To further illustrate the temporal consistency and dynamic evolution of this spatial attention across consecutive frames, a continuous visualization is provided in the supplementary video.

\section{Conclusions \& Future Work}
In this paper, we propose ColoACT, an autonomous navigation system that integrates an RGB-D-E-based Action Chunking Transformer policy (ColoACT policy) for a compact self-propelled Bevel-Gear-Based Endoscopic Robot (BGER). Trained by expert demonstrations, our policy inherits human navigation priors and reduces reliance on accurate modeling of viscoelastic, contact-rich interactions, mitigating sim-to-real issues. This method augments RGB with depth and a pseudo-elevation map to serve as a rich input for an ACT-based control policy, explicitly enhancing fold-ridge saliency and other high-frequency geometric cues. Leveraging action chunking and temporal ensembling, our approach achieves smooth, continuous motion for the proposed BGER while reducing reliance on manually engineered mode-switching and stepwise action prediction. Extensive \textit{ex-vivo} validation shows reliable performance and improved robustness, achieving success rates of 85.4\% and 72.5\% in straight and curved sections, respectively, with demonstrated adaptability to complex tortuous anatomies such as 90° turns, double-bend, and triple-bend colon. In the future, we plan to integrate force sensing to improve safety during tissue interaction and focus on validating the system in \textit{in-vivo} trials to address physiological dynamics such as peristalsis and varying lubrication, thereby improving practicality for future clinical evaluation.


%%%%%%%%%%%%%%%%%%%%%%%%%%%%%%%%%%%%%%%%%%%%%%%%%%%%%%%%%%%%%%%%%%%%%%%%%%%%%%%%

%%%%%%%%%%%%%%%%%%%%%%%%%%%%%%%%%%%%%%%%%%%%%%%%%%%%%%%%%%%%%%%%%%%%%%%%%%%%%%%%

%%%%%%%%%%%%%%%%%%%%%%%%%%%%%%%%%%%%%%%%%%%%%%%%%%%%%%%%%%%%%%%%%%%%%%%%%%%%%%%%

% \section*{ACKNOWLEDGMENT}

%%%%%%%%%%%%%%%%%%%%%%%%%%%%%%%%%%%%%%%%%%%%%%%%%%%%%%%%%%%%%%%%%%%%%%%%%%%%%%%%

\bibliographystyle{IEEEtran}
\bibliography{reference}
\end{document}